\documentclass[journal]{IEEEtran}  

\IEEEoverridecommandlockouts                              

\usepackage[pdftex]{graphicx}
\DeclareGraphicsExtensions{.pdf,.jpeg,.png}
\usepackage{amsmath} 
\usepackage{amssymb}  
\DeclareMathOperator{\E}{\mathbb{E}}
\DeclareMathOperator{\D}{\mathbb{D}}

\usepackage{tabularx,booktabs,multirow}
\usepackage{array}
\usepackage{float}
\usepackage{makecell}
\usepackage{tablefootnote}
\usepackage{multicol}
\usepackage{booktabs}
\usepackage{threeparttable}
\usepackage{url}
\usepackage[font=small,skip=10pt]{caption}
\usepackage[dvipsnames]{xcolor}
\usepackage{amsmath}
\usepackage{xcolor}
\usepackage{algorithmic}
\usepackage{dirtytalk}
\usepackage{hyperref}
\hypersetup{colorlinks,allcolors=red}
\usepackage[ruled,vlined]{algorithm2e}
\newcommand{\R}{\mathbb{R}}
\newcolumntype{P}[1]{>{\centering\arraybackslash}p{#1}}
\SetKwInOut{Input}{input}
\SetKwInOut{Output}{output}

\title{
Leveraging Kernelized Synergies on Shared Subspace for Precision Grasp and Dexterous Manipulation
}

\author{Sunny Katyara$^{1,2}$, Fanny Ficuciello$^{2}$,~\IEEEmembership{Senior Member,~IEEE}, Darwin Caldwell$^{1}$,~\IEEEmembership{Senior Member,~IEEE}, Bruno Siciliano$^{2}$,~\IEEEmembership{Fellow,~IEEE}, Fei Chen$^{1}$, ~\IEEEmembership{Member,~IEEE} 
\thanks{This research is supported by the project ``Improving Reproducibility in Learning Physical Manipulation Skills with Simulators Using Realistic Variations'' funded by EU H2020 ERA-Net Chist-Era program. \textit{(Corresponding author: Fei Chen)} }
\thanks{$^{1}$ Sunny Katyara, Darwin Caldwell, Fei Chen are with Active Perception and Robot Interactive Learning Laboratory, Department of Advanced Robotics, Istituto Italiano di Tecnologia, Via Morego 30, 16163, Genova, Italy (e-mail: {\tt\small name.surname@iit.it}).}
\thanks{$^{2}$ Fanny Ficuciello, Bruno Siciliano are with Department of Information Technology and Electrical Engineering and the Interdepartmental Center for Advances in Robotic Surgery, University of Naples Federico II, Naples 80125, Italy ({e-mail: \tt\small name.surname@unina.it}).}
}

\begin{document}

\maketitle
\IEEEdisplaynontitleabstractindextext
\IEEEpeerreviewmaketitle

\begin{abstract}

Manipulation in contrast to grasping is a trajectorial task that needs to use dexterous hands. Improving the dexterity of robot hands, increases the controller complexity and thus requires to use the concept of postural synergies. Inspired from postural synergies, this research proposes a new framework called kernelized synergies that focuses on the re-usability of the same subspace for precision grasping and dexterous manipulation. In this work, the computed subspace of postural synergies is parameterized by kernelized movement primitives to preserve its grasping and manipulation characteristics and allows its reuse for new objects. The grasp stability of the proposed framework is assessed with a force closure quality index. For performance evaluation, the proposed framework is tested on two different simulated robot hand models using the Syngrasp toolbox and experimentally, four complex grasping and manipulation tasks are performed and reported. The results confirm the hand agnostic approach of the proposed framework and its generalization to distinct objects irrespective of their shape and size.

\end{abstract}


\section{INTRODUCTION}

\IEEEPARstart{F}{or} more than three decades, roboticists have been actively trying to replicate the dexterity of the human hand and there have been excellent developments in the field of dexterous hands and advanced controls \cite{c1}\cite{c2}. But they have failed to achieve the universality and subtle behavior that characterizes dexterous manipulation, due to the limitations of available software and hardware. Building robot hands for dexterous manipulation that can emulate or even approach the functionality of humans is challenging owing to the fact that available sensors and actuators are not equivalent in size, precision and accuracy to the human muscles and skin. Dexterous manipulation epitomizes the ability of the hand to change the object pose from one configuration to another and can be achieved by re-grasping, finger gaiting and rolling/sliding \cite{c3}. 

   \begin{figure}[t]
      \centering
      \includegraphics[width=8.5cm]{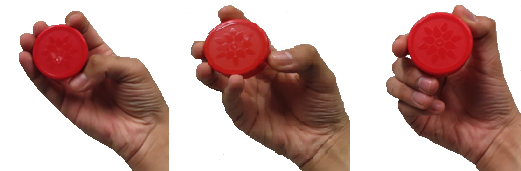}
      \caption{Representation of precision grasp and dexterous manipulation}
      \label{grasp_to_manipulation}
      \vspace{-15pt}
   \end{figure}
 
Dexterous manipulation by finger gaiting inherently depends upon the stability and configuration of the precision grasp to determine the minimum number and optimal position of the fingertips on the object's surface, as shown in Fig. \ref{grasp_to_manipulation}. However, this is an iterative process and to avoid computing optimal grasp each time, a database of grasps can be created and exploited to sample and rank the candidate grasps \cite{c4}\cite{c5}\cite{c6}. However, the grasp configurations do not only depend upon the object but the robot as well. With robots having increasing degrees of freedom, the controller complexity increases. To simplify the control of dexterous hands, inspiration is taken from the neuro-scientific behavior of the human brain, that suggests the use of postural synergies \cite{c7}\cite{c8}. The postural synergies form a reference subspace of coordinated human movements that are related to the hand kinematics. The postural synergies are computed from joint space configurations of the hand using statistical analysis methods such as:-, principle component analysis (PCA), expectation-maximization (EM), factor analysis etc.

Postural synergies computed on human hands require a mapping strategy to replicate the corresponding synergistic motions onto the kinematics of robot hands. There are three major mapping algorithms i.e, joint-joint mapping, Cartesian-space mapping and object-based mapping \cite{c9}\cite{c10}. However, all of these are prone to errors due to the dissimilarity in the kinematics and dimensions of the human and robot hands. Therefore, robot hands are either tele-operated during tasks or kinesthetically taught the required skills by a human expert \cite{c11}\cite{c12}. The statistical analysis can be now directly applied on the robot hand configurations to extract the corresponding synergy subspace. This direct interpretation of the postural synergy subspace is the most popular aspect in robot learning \cite{c13}.    

   \begin{figure}[tbp]
      \centering
      \includegraphics[width=8.5cm]{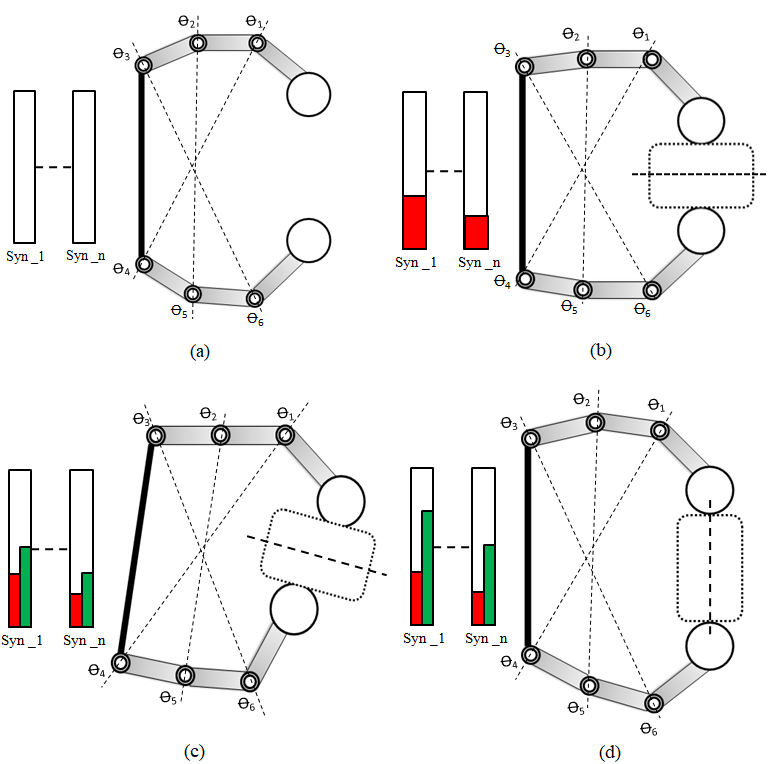}
      \caption{Conceptual representation of kernelized synergies on shared subspaces for grasping and manipulation, (a) represents the zero-offset pose of a hand with corresponding synergies in columns, (b) is the grasping pose obtained at respective values of synergies (red), (c) denotes the manipulation (continual rotation) of a grasped object by co-utilizing the grasping subspace (red) and manipulation parameters (green), (d) shows the quadrature rotation of grasped objects achieved with the combination of corresponding manipulation synergies (green). The dotted lines indicate the coordination among the hand joints in synergy subspace}
      \label{general_precision_manipulation}
      \vspace{-15pt}
   \end{figure}

Postural synergies have been effectively exploited for grasping with different designs of dexterous robot hands \cite{c14}\cite{c15}. But their application to dexterous manipulation is still challenging because it requires continuous coordination of all the fingers. Nonetheless, a few attempts have been made to use them for manipulation as well \cite{c16}\cite{c17} but generally these have not been effective as they require additional synergies to be used for different scenarios. This, of-course violates the prime motivation behind the use of postural synergies i.e, simplifying the robot hand control. Such limitations of postural synergies for dexterous manipulation confirm a marginal gap between the grasping and manipulation sub-spaces and thus require the introduction of middle-ware to link them. The possible intermediate link could be achieved through the use of a probabilistic kernel trick, which is an instance based learning approach rather than learning some fixed set of parameters \cite{c18}. The probabilistic kernel approach preserves the grasping and manipulation primitives of the synergy subspace and allows its reuse for gripping and manipulating daily life objects. This is the main motivation behind the development of our framework, which has been termed "Kernelized Synergies". The conceptual representation of kernelized synergies on a shared grasping and manipulation subspace is shown in Fig.\ref{general_precision_manipulation}. It shows how the same subspace may be shared between the grasping and manipulation components to execute the desired actions on the object.    

The rest of this paper is organized as follows; Section-II discuses recent research work in the field of postural synergies and our contributions. Section-III presents the research methodology applied to the formulation of kernelized synergies. Section-IV elaborates the grasp stability using the force closure property together with the soft synergy model for contact compliance. Section-V evaluates the performance of kernelized synergies when applied to two different simulated robot hand models using the SynGrasp toolbox.  Section-VI examines the experimental results obtained using kernelized synergies in four different complex scenarios and section-VII gives final conclusions about the development, application and outcome of proposed framework and its possible extensions for future work.  

\section{Related Works and Our Contributions}

For stable grasping and dexterous manipulation, a robot hand must exploit the object dimensions and environmental constraints and a method was introduced in \cite{c19}. It used tactile information with postural synergies to make a trade-off between the controller complexity and kinematic redundancy. This work was limited to the grasping of known objects in an unstructured environment. However, to automate and improve the synergy based grasping, a framework was discussed in \cite{c20}. This second framework was restricted to the synergistic grasping and did not examine manipulation. Hence, to cover synergy subspace also for manipulation, a technique was presented in \cite{c21} that employed additional synergies for manipulation but this increased the controller complexity and did not adapt to the dimensions of object. Therefore, to add adaptive characteristics to postural synergies and benefit from their smaller number, a new design and control architecture for robot hands was proposed in \cite{c22}. The developed architecture with only two synergies was able to perform grasping and manipulation for a large variety of objects. Yet, the framework was hand dependent as intelligence was embodied into its mechanical structure and was unable to comply with the system dynamics. So, to address the uncertainties related to the system dynamics during manipulation, a dynamic approach was presented in \cite{c23}. This approach used a kernel technique indirectly to estimate the variations in geometrical properties of the manipulated object but it suffers from dimensionality due to the inverse cubic dependence of its feature matrix. However, the promising potential of this concept motivated us to use a kernel trick on postural synergies for dexterous manipulation tasks.

\begin{table}[h]
\caption{Comparison among the existing state-of-art techniques and proposed framework.}
\label{characteristics}
\begin{center}
\begin{tabular}{P{5.5em}|P{4em}|P{5.5em}|P{4em}|P{4.5em}}
\hline
\textbf{Parameter} & \textbf{Soft synergies [18]} & \textbf{Manipulation synergies [14]} & \textbf{Complex synergies [10]} & \textbf{Our approach} \\
\hline
\textit{Task generalization} & {\checkmark} & {$\times$} & {$\times$} & {\checkmark} \\ 
\hline
\textit{Object adaptation} & {\checkmark} & {$\times$} & {$\times$} & {\checkmark} \\
\hline
\textit{Subspace retention } & {$\times$} & {\checkmark} & {$\times$} & {\checkmark} \\
\hline
\textit{Task prioritization} & {$\times$} & {$\times$} & {$\times$} & {\checkmark} \\
\hline
\textit{Properties preservation} & {\checkmark} & {$\times$} & {\checkmark} & {\checkmark}  \\
\hline
\textit{Grasping \& manipulation} & {$\times$} & {\checkmark} & {\checkmark} & {\checkmark}  \\
\hline
\end{tabular}
\end{center}

\begin{footnotesize}
\noindent Task generalization – perform tasks on new objects outside the data set\\ 
\noindent Object adaptation – adapt to size and shape of different objects\\
\noindent Subspace retention – same subspace used for grasping \& manipulation\\
\noindent Task prioritization – give preferences to tasks on their priority levels\\
\noindent Properties preservation -  preserve grasping properties of subspace\\
\noindent Grasping \& manipulation – able to grasping and then manipulation objects\\
\end{footnotesize}
\end{table}

To the best of our knowledge, there is no single control framework available to share the same subspace for precision grasping and dexterous manipulation and this is a key to the development of kernelized synergies. The major contributions of this research are; (1) to extract postural synergies for each training object and apply GMM-GMR to obtain a single synergistic trajectory, used to reproduce and generalize the taught hand postures (2) to use the kernelized movement primitives (KMP) to adapt to the unknown conditions and inherently kernelize the resultant subspace so that the learned skills are preserved globally and can be reused for new objects (3) to apply the force closure quality index together with a model of soft synergies to examine the grasp stability of kernelized synergies, (4) to evaluate the performance and hand agnostic approach of the proposed framework: this is at first tested on two different simulated robot hand models using the Syngrasp toolbox and then on a real setup with four complex tasks i.e, Pouring coffee and closing a jar, opening toolbox latches, grasping and manipulating two objects sequentially, and playing the board game \say{carrom}.    

To better understand the capabilities of the proposed method, a comparison is made with the state of art approaches used for synergistic grasping and manipulation in Table.\ref{characteristics}.


\section{Research Methodology}

\begin{figure}[t]
      \centering
      \includegraphics[width=8.5cm]{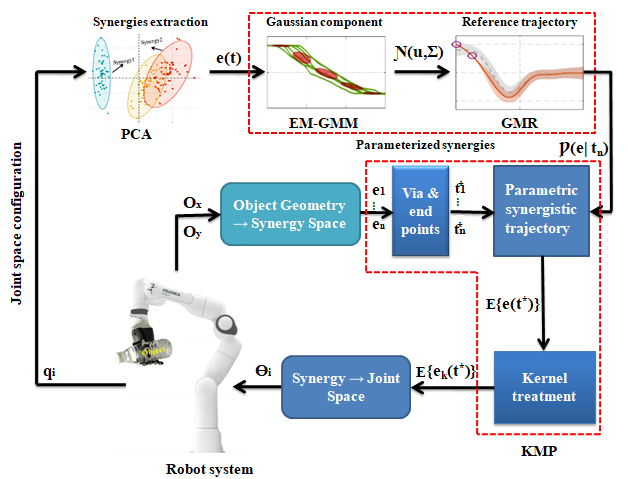}
      \caption{Block diagram of proposed methodology for the development of kernelized synergies.}
      \label{methodology}
      \vspace{-15pt}
   \end{figure}

Figure \ref{methodology} represents the block diagram of methodology proposed to perform the robust, reliable and adaptive grasping and manipulation with the dexterous robot hand. The methodology starts by teaching basic grasping and manipulation primitives to the robot hand using the training objects in Fig. \ref{objects} and recording the corresponding hand joint configurations. The hand joint configurations are normalized using PCA and the respective eigen vectors, so called postural synergies, are extracted. The coefficients of computed postural synergy evolve over the duration of the demonstration to obtain the corresponding synergistic trajectories. In order to account for inconsistencies during the demonstrations and obtain a generalized probabilistic trajectory, used to reproduce the learned skills , the Gaussian mixture model (GMM) is applied to approximate the distribution of the synergistic trajectories in terms of the Gaussian components. The expectation maximization (EM) initializes the Gaussian parameters i.e. weight, mean and co-variance. The Gaussian mixture regression (GMR) subsequently computes the conditional probability distribution of data in the GMM with respect to partially observed data during EM iterations and outputs the reference trajectory for demonstrated synergistic trajectories. In order to adapt to new conditions i.e, new objects, and encode the synergistic trajectories that exhibit time dependent variance, the KMP is exploited to generate a parametric synergistic trajectory. The use of kernel trick in KMP helps to preserve probabilistic properties (grasping and manipulation primitives) of parametric synergistic trajectory from multiple demonstrations, and alleviates the need of basis functions (features) used to estimate it. Geometrical information of the object is first translated into its corresponding synergistic data, which is then used to modulate the via points and end points of  the reference trajectory and thus the synergy subspace relearns the new task trajectory and with it develops the capacity to adapt to unknown conditions i.e. new objects. Finally, the mapping converts from the synergy subspace to joint space to perform the desired tasks on the robot hand.

\subsection{Extraction of Grasping and Manipulation Synergies}

\begin{figure}[t]
      \centering
      \includegraphics[width=8.5cm]{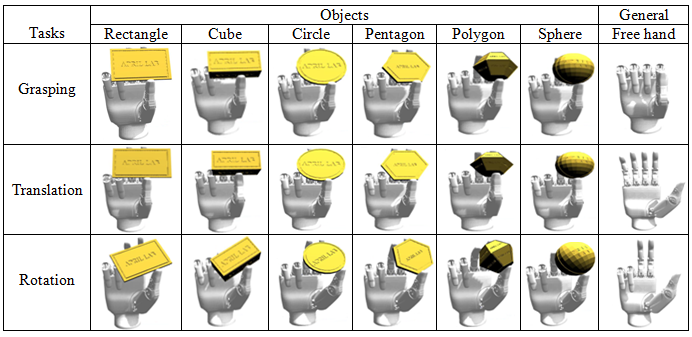}
      \caption{Training data set obtained by tele-operating the robot hand for grasp and manipulation primitives on given geometrical objects}
      \label{objects}
      \vspace{-15pt}
   \end{figure}

The postural synergies form a reduced subspace for coordinated motions of the robot fingers, whereby they can perform complex tasks such as; fine pre-shaping of fingers and dexterous manipulation. The postural synergies are the data driven control parameters and are obtained from the statistical analysis (PCA) of the hand joint configurations. The data is generated by tele-operating the robot hand to perform a range of precision grasping and dexterous manipulation (rotation and translation) actions on a series of geometrical objects as shown in Fig. \ref{objects}. The robot hand is taught basic grasping and manipulation (rotation and translation) primitives so that the subspace can be generalized to perform different task on common objects used in daily life. 
   
Let the hand joint vector be denoted by $\theta_k$ at the $k^{th}$ training task then the nominal posture of the robot hand for $K$ trials is determined by Eq. \ref{eqt_2}
   
\begin{equation}
s_0={\frac{1}{K}}{\sum_{k=1}^{K}}{\theta_k}
\label{eqt_2}
\end{equation} 

The mean position of the robot hand is obtained by subtracting the nominal posture from its current joint configuration ${\hat{\theta_k}}$=$\theta_k$-$s_0$ and then concatenating them into a row vector as $C$=$[{\hat{\theta_1}}......{\hat{\theta_K}}]^{T}$. The PCA is applied on configuration matrix $C$ and reduced subspace of postural synergies $\hat{E}$ is obtained \cite{c24}.

To reproduce the required posture on the robot hand for a given object to be grasped and manipulated, a proper choice of postural synergy co-efficient $(e_i)={e_{g1}+e_{m1},e_{g2}+e_{m2},...e_{gn}+e_{mn}}$ should be selected. The corresponding synergy co-efficient is determined by Eq. \ref{eqt_5}

\begin{equation}
{e_i}={\hat{E}}^{\dagger}{({\theta_i}-{s_0})}
\label{eqt_5}
\end{equation}   

The matrix ${\hat{E}}^{\dagger}$ denotes the pseudo inverse of the reduced synergy matrix and thus the projection of each robot hand configuration on the synergy subspace is defined as ${{\hat{\theta}}_i}={s_0}+{\hat{E}}{e_i}$

\begin{figure*}[t]
      \centering
      \includegraphics[height=3.5cm, width=15cm]{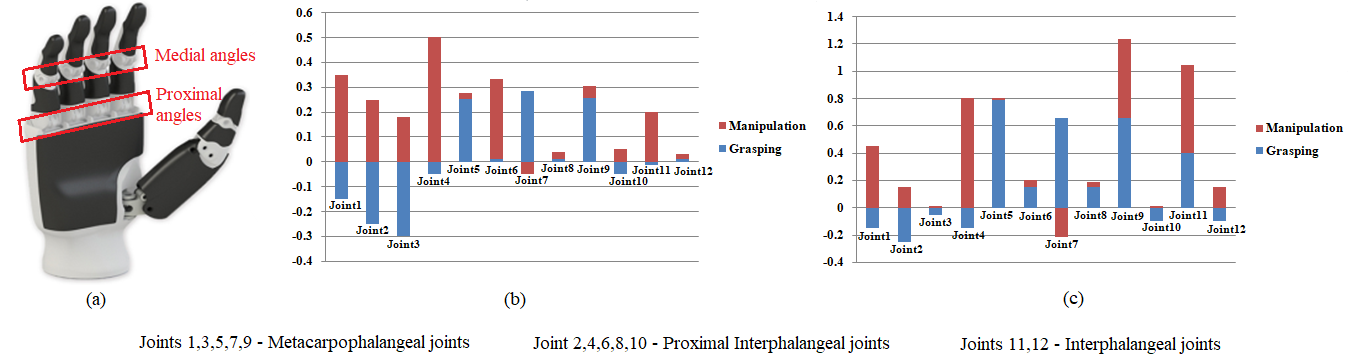}
      \caption{Graphical representation of synergy subspace for 6 DoF robot hand, (a) is the fully actuated INSPIRE robot hand, (b) and (c) represent the first and second synergies composed of both grasping and manipulation components respectively}
      \label{synergies}
      \vspace{-15pt}
   \end{figure*}

The subspace of two predominant postural (grasping and manipulation) synergies for a 6 DOF INSPIRE robot hand having 12 joints in  Fig. \ref{synergies} (a) \cite{c25}, computed using training objects in Fig. \ref{objects} is graphical shown in Fig. \ref{synergies} (b and c). It can be seen that the first synergy controls the proximal and medial angles of all the finger joints while the second synergy mainly regulates the movement of the thumb and index finger for the respective grasping and manipulation tasks. Moreover, the relative positions of the thumb and index finger help in moving from one grasping posture to other while manipulating an object. 

\subsection{Parameterization and Kernelization of Postural Synergies}

The postural synergies determined in the previous section for each training object in Fig. \ref{objects} are evolved over the duration of demonstrations to obtain corresponding synergistic trajectories. In order to capture the joint probabilistic distributions among these synergistic trajectories $e(t)$, the GMM is applied and is defined by Eq. \ref{eqt_13} \cite{c26}\cite{c27}

\begin{equation}
    \begin{bmatrix} {t} \\ {e} \end{bmatrix} {\sim}{\sum_{n=1}^{N}}{\pi_n}{\mathcal{N}}({{\mu}_n},{{\Sigma}_{n}})
\label{eqt_13}
\end{equation}

\begin{algorithm}
\SetAlgoLined
\Input{${q_i}\longrightarrow{\text{joint hand configurations}}$}   
\Output{${\rho}(e(t))\backsim\text{function}(\E(e(t),\D(e(t)))$}
 \While{$({q_0}\in{q_i})\longrightarrow{C}$}{
 $function (C)\longrightarrow(E,e)$\;
  \eIf{$e\in(E)>0.85$}{
   ${\hat{E}}$\;
   }{
   $\phi{\longrightarrow}{\text{null matrix}}$\;
  }
 }
\caption{Kernelized synergies}
\ForEach{$e(t_n)\backsim{\mathcal{N}}({\mu_n},{\Sigma_n})$}{
 $function (\pi_n,\mu_n,\Sigma_n)$\;
  \eIf{${\Upsilon{\neq}0}\longrightarrow{\text{priority exist}}$}{
   $\rho(e(t))\backsim{\prod}{\mathcal{N}}(\mu_w,{\Sigma_w/\Upsilon})$\;
   
   }{
   $\rho(e(t))\backsim{O_{min}}(\mu_w,{\Sigma_w})$\;
  }
}
\end{algorithm}

The terms ${\pi}_n$,${\Sigma}_n$,${\mu}_n$ represent the prior probability, covariance and mean of the $n^{th}$ Gaussian component respectively and N indicates the number of Gaussian components used to approximate the given synergistic trajectories.  Moreover, in order to obtain a reference synergistic trajectory ${({\hat{e}})_{n}}^{N}=1$ to be followed by the hand to reproduce the demonstrations, the GMR is applied. This actually describes the conditional joint probability distribution of the GMM i.e, ${e_n}|t{\sim}{\mathcal{N}}({{\mu}_n},{{\Sigma}_n})$. The conditional mean and covariance of the reference synergistic trajectory are thus computed by Eq. \ref{eqt_14}, where ${\mu_n}^{t}$ and ${\Sigma_n}^{tt}$ represent instantaneous mean and co-variance respectively

\begin{equation}
\!
\begin{aligned}
{\hat{\mu}}_{n}= {\sum_{n=1}^{N}}{\frac{{\pi}_n{\mathcal{N}({}t_n}|{{\mu}_n}^{t}{{\Sigma}_n}^{tt})}{{\sum_{n=1}^{N}}{\mathcal{N}({}t_n}|{{\mu}_n}^{t}{{\Sigma}_n}^{tt})}}{{\mu}_n}^{e}+{\vartheta} \\
{\hat{\Sigma}}_{n}= {\sum_{n=1}^{N}}{\frac{{\pi}_n{\mathcal{N}({}t_n}|{{\mu}_n}^{t}{{\Sigma}_n}^{tt})}{{\sum_{n=1}^{N}}{\mathcal{N}({}t_n}|{{\mu}_n}^{t}{{\Sigma}_n}^{tt})}}{\varphi}+{\varepsilon}
\end{aligned}\label{eqt_14}
\end{equation}

where;

\begin{footnotesize}
\noindent ${\vartheta}={{\sum}_n}^{et}({{\sum}_n}^{tt})^{-1}({t_n}-{{\mu}_n]^{t}}$\\ 
\noindent ${\varphi}=({{\sum}_n}^{ee}-{{\sum}_n}^{et}({{\sum}_n}^{tt})^{-1}{{{\sum}_n}^{te}})$\\
\noindent ${\varepsilon}=({{\mu}_n}^{e}+{{\sum}_n}^{et}({{\sum}_n}^{tt})^{-1}({t_n}-{{\mu}_n}^{t})({{\sum}_n}^{tt})^{-1}({{t_n}-{\mu}_n}^{t})^{T}-(\hat{{\mu}_n})(\hat{{\mu}_n})^{T}$\\
\end{footnotesize}

To generalize the learned synergy subspace to wider set of new objects, the KMP \cite{c28} is exploited to generate the parametric synergistic trajectory and is defined by the Eq. \ref{eqt_17}, where ${\mu}_w$ and ${\Sigma}_w$ are weighted mean and co-variance respectively.

\begin{equation}
    {e(t)}{\sim}{\mathcal{N}}({\Theta}(t)^{T}{{\mu}_w},{\Theta}(t)^{T}{{\Sigma}_w}{\Theta}(t))
\label{eqt_17}
\end{equation}

The goal of parametric synergistic trajectory in Eq. \ref{eqt_17} is to follow the reference synergistic trajectory in Eq. \ref{eqt_14} and to do so, the Kullback-Leibler (KL) divergence criteria is applied to minimize the distance between two i.e, $O_{min}(\mu_w,\Sigma_w)$. It eventually determines the weighted mean and co-variance in Eq. \ref{eqt_17}. Hence, the optimal optimal values of the weighted mean and co-variance are found by Eq. \ref{eqt_21} \cite{c28}, with $\lambda$ being regularization parameter, $\Phi$ is a matrix of basis functions (features), ${\mu}=[{\hat{\mu}_1},{\hat{\mu}_2},.....{\hat{\mu}_N}]$ and ${\Sigma}={blockdiag}({\hat{\Sigma}_1},{\hat{\Sigma}_2},.....{\hat{\Sigma}_N})$ represent the mean and co-variance matrices of demonstrated synergistic trajectories respectively. 

\begin{equation}
\begin{gathered}
{{\mu}_w}={\Phi}({\Phi}^{T}{\Phi}+{\lambda}{\Sigma})^{-1}{\mu} \\ 
{{\Sigma}_w}=N({\Phi}{{\Sigma}^{-1}}{{\Phi}^{T}}+{\lambda}{I})^{-1}
\end{gathered}
\label{eqt_21}
\end{equation}

In order to preserve the probabilistic properties of grasping and manipulation synergies so that they can be reused for new objects, and also to alleviate the use of basis functions (features) whose number increases exponentially for complex tasks, the Eq. \ref{eqt_21} is kernelized in KMP. Kernel treatment of basis functions (features) used to encode the synergistic trajectory is defined as $k({t_i},{t_j})={\phi}(t_i),{\phi}(t_j)$.

Therefore, for any new input $t^{*}$ (new shape or/and size of object), the expected values of the mean and co-variance of synergistic component are determined by Eq. \ref{eqt_22} \cite{c28}, where $k^{*}$ represents the kernel function for the new object input and is defined as ${{k}^{*}}=[{k}(t^{*},{t}_{1}),{k}(t^{*},{t}_{2}),.....{k}(t^{*},{t}_{N})]$ and $K$ is the kernel matrix.

\begin{equation}
\begin{gathered}
{\E}(e(t^{*}))={k}^{*}({K}+{\lambda}{I})^{-1}{\mu} \\
\D(e(t^{*}))={\frac{N}{\lambda}}({k}({t}^{*},{t}^{*})-{k}^{*}({K}+{\lambda}{\Sigma})^{-1}k^{*T})
\end{gathered}
\label{eqt_22}
\end{equation}

The frame of kernelized synergies can also prioritize the different grasping and manipulation tasks on the basis of their assigned weights in the joint probability distribution. For a set of M reference synergistic trajectory distributions, the corresponding synergistic input $\{{t_n},{\hat{e}}_{n,m}\}$ assigned with priority $\Upsilon_{n,m} {\in} (0,1)$ together represented as ${\{{\{{{t_n},{\hat{e}}_{n,m}},\Upsilon_{n,m}\}_{n=1}^{N}}\}_{m=1}^{M}}$ should satisfy the condition ${\sum_{m=1}^{M}}\Upsilon_{n,m}=1$. Therefore, prioritizing the tasks in kernelized synergy subspace actually  corresponds  to  the  product  of  its M normal distributions with $m= 1,2,...M$ and is thus explained by Eq.\ref{eqt_31} \cite{c28}  

\begin{equation}
\mathcal{N}({\mu_n}^{T},{\Sigma_n}^{T}) {\propto}{\prod_{m=1}^{M}}\mathcal{N}({\hat{\mu}_{n,m}},{\hat{\Sigma}_{n,m}}/\Upsilon_{n,m})
\label{eqt_31}
\end{equation}

The complete methodology for kernelized synergies framework is summarized in Algorithm 1.

\subsection{Object geometry to synergies transformation}

\begin{figure}[t]
      \centering
      \includegraphics[width=8.5cm]{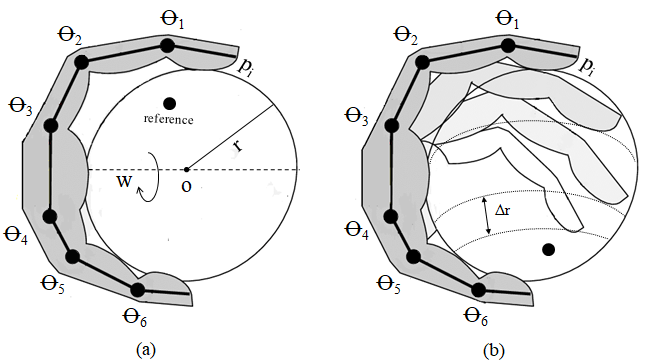}
      \caption{Transformation from task space into synergistic subspace, (a) denotes the translational and angular movement indentations for object grasped by a robot hand, (b) illustrates the possible translation and orientation of the object as it is being manipulated by the hand.}
      \label{manipulate}
      \vspace{-15pt}
   \end{figure}

The geometry of different objects acts as an environmental descriptor i.e, via-points and end-points for the reference synergistic trajectory as described in the previous section. Since, the dimensions of the object are in real coordinates and need to be transformed into respective synergistic values. The hard finger contact model \cite{c29} is considered for the robot-object interaction. According to Fig. \ref{manipulate}, the relationship between the motion of an object and the contact points of robot hand manipulating it is defined by Eq. \ref{eqt_36} \cite{c30}
   
\begin{equation}
{\dot{p}}= {A_{m}}\begin{bmatrix} {\dot{o}} \\ {\omega} \\ {\dot{r}}  \end{bmatrix}
\label{eqt_36}
\end{equation}

Where ${\dot{o}}$ and ${\omega}$ represent the linear and angular velocities of the object and ${\dot{r}}$ denotes the displacement of object with reference to the contact points $p$. The quantity ${A_{m}}$ is called the motion transfer matrix and is defined by Eq. \ref{eqt_37}

\begin{equation}
{A_{m}}= \begin{bmatrix} I & -{[p_{1}-{o}]_{x}} & {(p_{1}-{o})} \\ \cdots & \cdots & \cdots \\ I & -{[p_{i}-{o}]_{x}} & {(p_{i}-{o})} \\ \cdots & \cdots & \cdots \end{bmatrix}
\label{eqt_37}
\end{equation}

In order to determine the synergistic values, the joint configuration of the robot hand is required. The relation between the robot hand contact points and its joint configuration is defined by the standard differential kinematics $\dot{p}=J_{h}{\dot{\theta}}$ where $J_{h} {\in} {\R}^{{n_c}{\times}{n_q}}$ is the hand Jacobian matrix with $n_c$ and $n_q$ being the number of hand contact points and joints respectively. Therefore, the mapping from hand joint velocities to the corresponding synergistic velocities is given by Eq. \ref{eqt_38}

\begin{equation}
    {\dot{e}}={J_h}{{A_m}^{\dagger}}{C_h}{S^{\dagger}}{\dot{\theta}}
\label{eqt_38}
\end{equation}

Where $C_h$ is the hand joint compliance matrix, $S$ is the synergy matrix of the hand and ${\dagger}$ represents the pseudo inverse of the respective quantities.  

\subsection{Synergy to Joint Space Mapping}
Once the kernelized synergies for a particular task are determined, the next step is to command the robot hand to execute it. In order to comply with hand joint control, the mapping is performed from the synergistic subspace to the joint space using a relation defined by Eq. \ref{eqt_39} 

\begin{equation}
    {\theta}={S}{e}+{\theta_0}
\label{eqt_39}
\end{equation}

The matrix $S$ represents the postural synergy sub-spaces of the robot hand, $e$ denotes the corresponding synergistic coefficients and $\theta_0$ indicates the initial hand joint configuration. It must be noted that the dimensions of $S$ depend upon the number of postural synergies considered to approximate the respective grasping and manipulation configurations.

\section{Grasp Stability Analysis}

The desired contact forces are not only required to hold the object in a stable position but also to avoid damage to the object or robot itself. Since, the joint torque of a DC motor is related to its current values by ${\tau}={K_m}{I}^2$ where $K_m$ is the motor design constant. According to differential kinematics, the relationship between contact forces and joint torques is defined by $f_c={J_h}{\tau}$. Therefore, there exist a kineto-static duality between the synergistic forces and velocities and hence the force synergies ${\eta}{\in}{\R}^{n_s}$ are defined by Eq. \ref{eqt_40}, with $n_s$ being the number of synergy components considered 

\begin{equation}
    {\eta}={S}^T{J_h}^T{f_c}
\label{eqt_40}
\end{equation}

With the model in Eq. \ref{eqt_40}, the co-efficients of kernelized synergies are modulated until the steady state contact forces are obtained. This must occur within the defined threshold on motor current to ensure the grasp stability. Moreover, the force closure property is considered to evaluate the grasp quality of the proposed framework \cite{c31}. The problem is formulated as a cost function $\Gamma$ minimization to ensure the stability of the grasp. Since, the general solution to the problem of contact forces balancing the grasped object is described by Eq. \ref{eqt_41}

\begin{equation}
    {f_c}={G}^{\dagger}{\omega}+{\xi}{\Delta}{q_{ref}}
\label{eqt_41}
\end{equation}

Where ${f_c}{\in}{\R}^{3n_c}$ is the vector of contact forces applied to $n_c$ contact points, ${G}^{\dagger}$ is the pseudo-inverse of grasp matrix, ${\omega}{\in}{\R}^{6}$ is the vector of external spatial forces, ${\xi}$ is the subspace of internal forces and ${\Delta}{q_{ref}}$ is the change in joint reference position. The matrix $\xi$ maps the joint positions to the internal forces generated during the hand-object interaction. Now, the change in the joint positions can be defined in synergistic subspace by ${\Delta}{q}={S}{\Delta}{e}$ and ${\Delta}{q}={\Delta}{q_{ref}}-{C}{\Delta}{\tau}$, which is the model of soft synergy. The grasping problem can now be defined in terms of the synergistic control framework by Eq. \ref{eqt_42}    

\begin{equation}
    {f_c}={G}^{\dagger}{\omega}+{\xi}{\Delta}{e}
\label{eqt_42}
\end{equation}

With reference to Eq. \ref{eqt_42} and exploiting Coulomb's friction cone criteria, the synergy-based grasp quality matrix using force closure property is defined as the minimization of the cost function ${\Gamma}({\Delta}{e})$ with respect to ${\Delta}{e}$. 
Let ${\Omega}_{i,j}^{p}{\in}{{\R}^{k}}$ represents the set of grasp variables $g$ that fulfill the friction cone constraints with small positive margin $p$, where $k$ is the dimension of ${\xi}$. Therefore, for $i-th$ contact and $j-th$ constraint, ${\Gamma}$ is found to be summation of respective terms:${\Gamma}({\omega},{g})={{\sum}_i}{{\sum}_j}{\Gamma}_{i,j}({\omega},{g})$ and defined by Eq. \ref{eqt_43} \cite{c32}

\begin{equation}
    {\Gamma}_{i,j}={\begin{cases} 
      ({2}{{\sigma}_{i,j}^{2}}({\omega},{g}))^{-1} & g\in {\Omega}_{i,j}^{p} \\
      {a}{{\sigma}_{i,j}^{2}}({\omega},{g})+{b}{\sigma}_{i,j}({\omega},{g})+c & g\notin {\Omega}_{i,j}^{p}
   \end{cases}}
\label{eqt_43}
\end{equation}

Where a, b and c are the constants conditioned by the properties of ${\Gamma}$. The cost function ${\Gamma}$ represents the quality of grasp, since its inverse indicates the grasp margin from violating the friction cone constraints and is used for planning the dexterous manipulation. 

Therefore, to evaluate the force closure property of kernelized synergies, the second correction term from Eq. \ref{eqt_43} is included in the desired reference synergistic trajectory so that the movements of the fingers are against the gradient of ${\Gamma}$. The main goal is to reproduce the synergistic movements that minimize the cost function and this is formulated by Eq. \ref{eqt_44} with ${\kappa_q}<{0}$ being a constant gain chosen experimentally according to the system dynamics.    

\begin{equation}
    {\Delta}{e}={\kappa}_{q}{\frac{\partial {\Gamma}}{\partial {e}}}{\Delta}{t}
\label{eqt_44}
\end{equation}

\section{Simulation Analysis}

\begin{figure}[tbp]
      \centering
      \includegraphics[width=8.5cm]{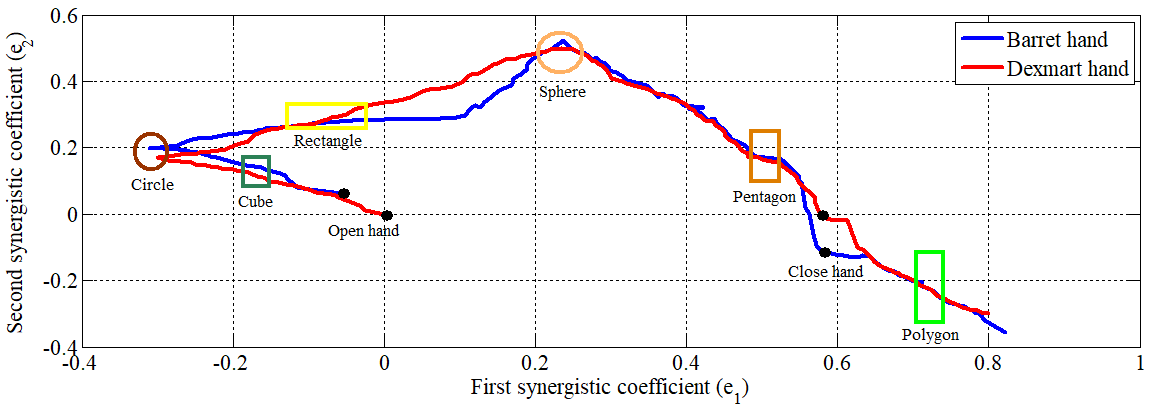}
      \caption{Training trajectories of two principal synergistic coefficients for the Barrett and Dexmart hands using given geometrical objects}
      \label{training_trajectory}
      \vspace{-15pt}
   \end{figure}

The proposed framework of kernelized synergies is numerically evaluated on two different robot hand models in the SynGrasp MATLAB toolbox \cite{c33}. The first model is an anthropomorphic under-actuated Dexmart hand having 20 DoF with size and kinematics structure very similar to a human hand \cite{c34}. Another model is the Barrett Hand, which is a three fingered fully actuated robot hand  with 8 joints and 4 DOFs \cite{c35}. Both the robot hand models are trained using the same geometrical objects as shown in Fig. \ref{objects} and the corresponding trajectories of the synergistic coefficients are reported in Fig. \ref{training_trajectory}. The simulations are performed with MATALAB2019b on a 2.6-GHZ Intel Core i5, 8-GB RAM computer.

\begin{figure*}[h!]
      \centering
      \includegraphics[height=7cm, width=18cm]{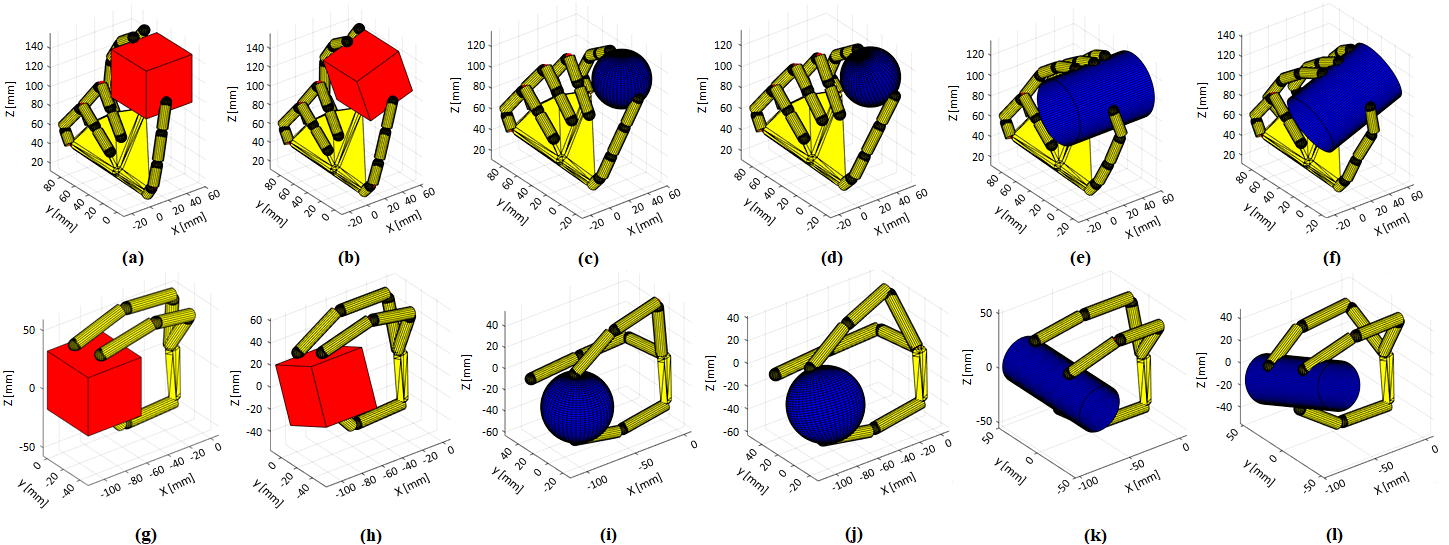}
      \caption{Simulation analysis of kernelized synergies on two different robot hands. In the first row (a-f), the Dexmart hand is grasping and manipulating (rotating) three distinct objects of different shapes and sizes and in the second row (g-l), the Barrett hand is performing similar actions on the given objects but due to its limited dexterity it only exhibits two different postures i.e, bipodal and tripodal except quadpodal}
      \label{simulink_analysis}
      \vspace{-10pt}
   \end{figure*}

To examine the application of kernelized synergies on the given robot hand models, precision grasping and manipulation (rotation) of three distinct objects, of different sizes i.e, cube, sphere and cylinder were considered, as shown in Fig. \ref{simulink_analysis}. Due to the compliance of the fingers introduced by the kernelized synergies, the contact forces between the object and hand are compensated by using hard finger contact model for both hands according to Eq. \ref{eqt_42}. It can be seen in Fig. \ref{simulink_analysis} that both the hands are able to adapt to the varying dimensions of objects and also maintain the grasp stability according to the force closure optimization criteria Eq. \ref{eqt_44}. The quality index computed for all the different grasps with both the hands are reported in Table. \ref{grasp_quality}. The reduced values of the quality index in case of the Dexmart hand confirm its better performance when stably grasping and manipulating objects due to its higher dexterity as compared to Barrett hand.

\begin{figure*}[t]
      \centering
      \includegraphics[height=8cm, width=18cm]{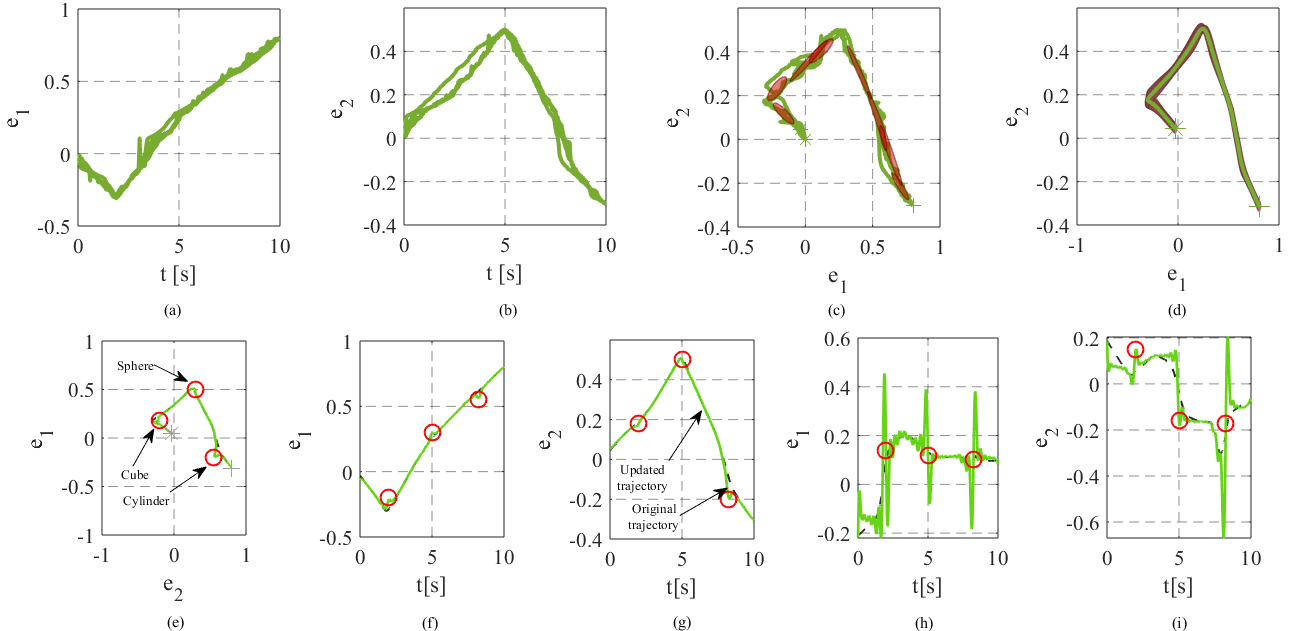}
      \caption{Kernelized synergistic profile of the Dexmart hand when grasping and manipulating three objects. (a and b) represent the training trajectories of the first two synergistic coefficients for the defined data-set, (c) is the relative trajectory approximated with Gaussian components shown in red ellipses , (d) is the final reference trajectory generated by the GMR with mean value shown by the green curve and the range of its variance indicated by shaded red area, (e) is the updated reference trajectory for grasping and manipulation (rotation) of three distinct objects, (f-g) are the parametric synergistic trajectories reproduced over the set of three via-points (for three objects), (h-i) show the corresponding synergistic velocities used to speed up or slow the hand posture adaption }
      \label{kernalized_synergy_profile}
      \vspace{-10pt}
   \end{figure*}

\begin{table}[t]
\caption{Quantitative computation of force closure quality index for both robot hand models.}
\label{grasp_quality}
\begin{center}
\begin{tabular}{P{5.5em}|P{5.5em}|P{5.5em}|P{6.5em}}
\hline
\textbf{Hand} & \textbf{Object} & \textbf{Grasp} & \textbf{Quality Index Gradient} \\
\hline
\textit{} & {Cube} & {Tripodal} & {$2.3\times10^7$} \\ 
\textit{Barret} & {Sphere} & {Bipodal} & {$3.7\times10^6$} \\
\textit{} & {Cylinder} & {Tripodal} & {$4.1\times10^7$} \\ 
\hline
\textit{} & {Cube} & {Tripodal} & {$1.7\times10^6$} \\ 
\textit{Dexmart} & {Sphere} & {Bipodal} & {$6.5\times10^5$} \\
\textit{} & {Cylinder} & {Quadpodal} & {$2.2\times10^7$} \\ 
\hline
\end{tabular}
\end{center}
\end{table}

The kernelized synergistic profile of the Dexmart hand grasping and manipulating the given objects is presented in Fig. \ref{kernalized_synergy_profile}. Figs. \ref{kernalized_synergy_profile} (a and b) denote the trajectories of two synergistic coefficients computed with Eq. \ref{eqt_5} over the total duration of demonstrations. In order to find the correlation among such synergistic trajectories, the GMM is applied according to Eq. \ref{eqt_13} and is illustrated in Fig. \ref{kernalized_synergy_profile} (c), where the red ellipses indicate the corresponding Gaussian components approximating the synergistic trajectories. For such probabilistic trajectory to achieve the desired poses under the different conditions, the reference trajectory in Fig. \ref{kernalized_synergy_profile} (d) is calculated using Eq. \ref{eqt_14}. For any new object, the proposed framework updates the via points and end points of the reference trajectory pursuant to Eq. \ref{eqt_22} such that it can be grasped and manipulated as shown in Fig. \ref{kernalized_synergy_profile} (e). Figs. \ref{kernalized_synergy_profile} (f and g) represent the updated trajectories of two synergistic components for the given three objects and their corresponding synergistic velocity profiles in Figs. \ref{kernalized_synergy_profile} (h and i) with desired via-points being marked. It is evident from the synergistic velocities that the hand experiences an overshoot for the short duration when it breaks contact with objects. This is obvious during the transition from contact state to non-contact configuration. 

\section{Results And Discussion}

To evaluate the performance of proposed framework, four distinct complex grasping and manipulation tasks are considered. All the given tasks are still based on basic grasping and manipulation primitives taught to the robot hand. 
   
   \begin{figure*}[h]
      \centering
      \includegraphics[height=3.5cm, width=18cm]{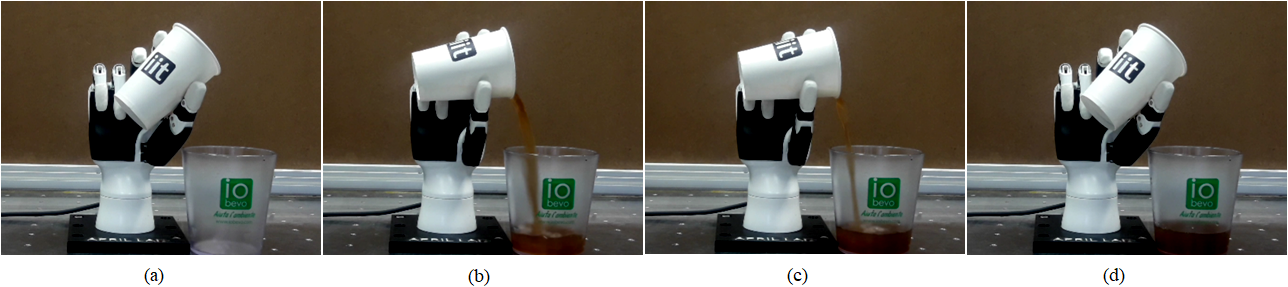}
      \caption{A robot hand pouring coffee from paper cup into a glass (a) illustrates the initial configuration of the hand grasping the coffee cup and the empty glass on the table, (b and c) show the flow of coffee into the glass at a rate determined by the relative movements of the fingers, (d) denotes the reoriented pose of coffee cup to stop the flow of coffee into the glass}
      \label{coffee}
      \vspace{-10pt}
   \end{figure*}
   
   \begin{figure*}[h]
      \centering
      \includegraphics[height=3.5cm, width=18cm]{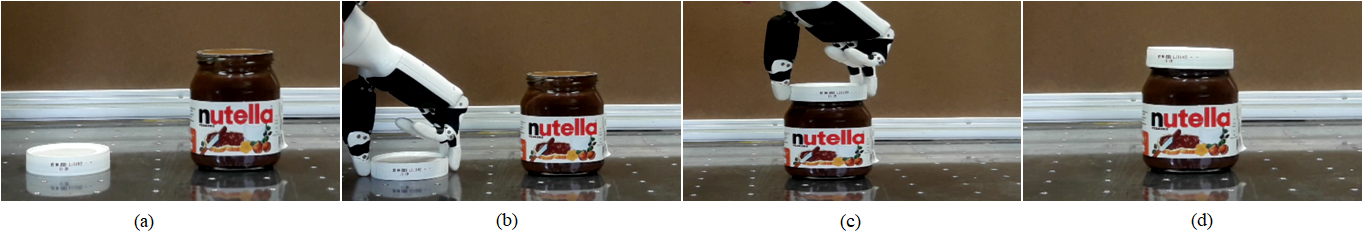}
      \caption{A robot hand closing the nutella jar, (a) illustrates the nuttela jar and its lid, (b) indicates robot hand grasping the lid in tripod (3 fingered) precision grasp, (c) shows the scenario where the lid is placed on the jar while maintaining a stable grasp, (d) displays the final pose of jar with the lid screwed onto it. }
      \label{nutella}
      \vspace{-10pt}
   \end{figure*}
   
   \begin{figure*}[t]
      \centering
      \includegraphics[height=3.5cm, width=18cm]{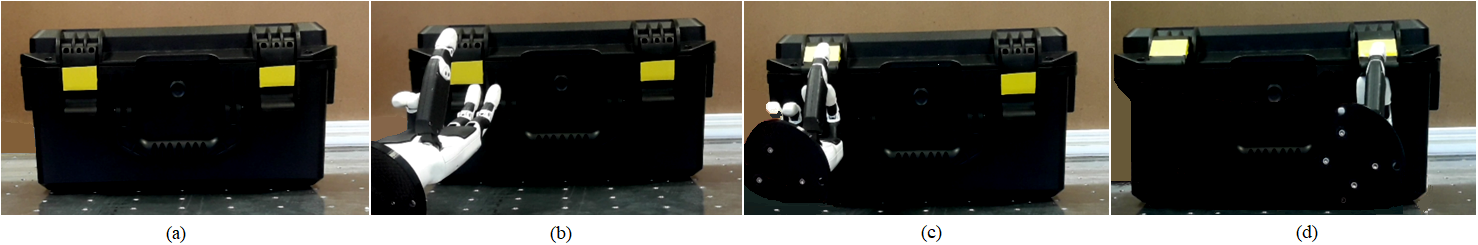}
      \caption{A Robot hand opening a toolbox using its learned translation manipulation primitive from the kernelized synergy subspace, (a) indicates the toolbox with both its latches closed initially, (b) represents the scenario in which the robot hand assumes a tripod (3 fingered) pose with its little and ring fingers closed, (c) illustrates the robot hand opening the first latch of toolbox with its index and middle fingers, (d) shows similar robot hand action to open the second latch}
      \label{box}
      \vspace{-10pt}
   \end{figure*}
   
   \begin{figure*}[t]
      \centering
      \includegraphics[height=4cm, width=18cm]{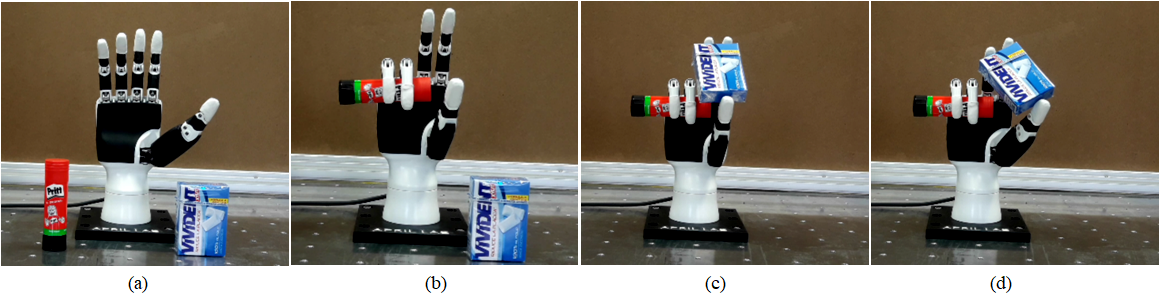}
      \caption{A robot hand controls its two parts independently by assigning priories to the synergistic components to grasp two different objects sequentially and manipulating one of them, (a) is the open hand configuration of the robot hand, (b) shows the grasping of the first object with the little and ring fingers, (c) is the sequential grasping of the second object with the thumb, index and middle fingers in a tripod configuration, (d) illustrates the clockwise orientation of the second object using its manipulation rotation primitive }
      \label{priority}
      \vspace{-10pt}
   \end{figure*}

   \begin{figure*}[t]
      \centering
      \includegraphics[height=3cm, width=18cm]{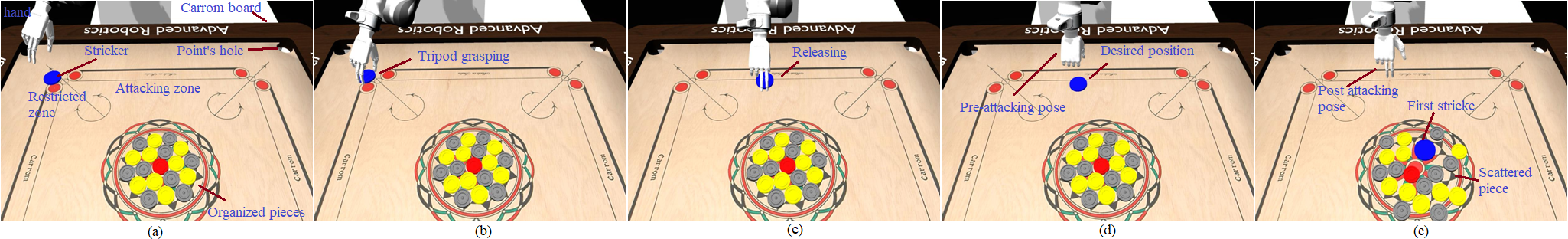}
      \caption{Robot playing carrom board game with the proposed framework, (a) denotes the initial configuration of the hand and striker which is in restricted zone, (b) shows the hand grasping striker with its index, middle and thumb in one of its taught precision posture, (c) reports the pose of hand releasing the striker at desired position in attacking zone, (d) represents the pose of hand which is stretched in to develop the required pushing force, (e) illustrates the pushing action of hand on the striker which leads it to the center to spread the pieces accordingly}
      \label{carrom_game}
      \vspace{-10pt}
   \end{figure*}
   
For task1. the robot hand utilizes its rotation manipulation primitive to pour the coffee from the cup into the glass and to close the nutella jar as presented in Fig. \ref{coffee} and Fig. \ref{nutella} respectively. For the coffee pouring task, the robot hand holding the paper coffee cup with ${e_{g1}=0.46, e_{g2}= 0.17}$ is required to pour the coffee into empty glass on the table in Fig. \ref{coffee} (a). The paper coffee cup is oriented with corresponding manipulation coefficients $e_{m1}=0.47$ to $0.54$, $e_{m2}= 0.18$ to $0.27$ and the coffee start to flow into a glass gradually depending upon the degree of orientation of fingers as demonstrated in Figs. \ref{coffee} (b) and (c). Finally, the cup is rotated backwards with $e_{m1}=0.54$ to $0.47$, $e_{m2}= 0.27$ to $0.18$ so that only the required amount of coffee is poured into a glass as displayed in Fig. \ref{coffee} (d). When closing the jar in Fig. \ref{nutella} (a), the robot hand first grasps the lid with the corresponding grasping components ${e_{g1}=-0.27, e_{g2}= 0.16}$ and then places it on top of the jar at the desired position as shown in Figs. \ref{nutella} (b) and (c) respectively. Finally, the robot hand turns the lid in a clockwise direction at $e_{m1}=-0.31$ to $-0.39$, $e_{m2}= 0.17$ to $0.21$ such that it is screwed onto the jar as illustrated in Fig. \ref{nutella} (d). It must be noted that both the tasks are performed primarily with the fingers without any compensation at the wrist level.     
   
During task2, the robot hand uses its translation manipulation primitive to open the toolbox latches with the required pulling action as shown in Fig. \ref{box}. Fig. \ref{box} (a) illustrates the toolbox with latches that are initially closed as marked yellow and the robot hand first assumes the desired pose at ${e_{g1}=-0.12, e_{g2}= 0.25}$ and then opens the first and second latches with its index and middle fingers at $e_{m1}=0.16$ to $0.23$, $e_{m2}= -0.17$ to $-0.05$ by utilizing its translation primitive to generate the corresponding outward force as displayed in Figs. \ref{box} (b), (c) and (d) respectively. Due to the compliance introduced by the kernelized synergies, the interaction between the fingers-tips and the latches is modulated from 0.85 N/sec to 1.23 N/sec during the pulling action according to Eq. \ref{eqt_42}.      
   
Task3 is about assigning priorities to the synergistic components according to Eq. \ref{eqt_31} to grasp two objects sequentially and then manipulate one of them while maintaining a stable grasp as shown in Fig. \ref{priority}. In this case, the kernelized synergies exploit their priority characteristics to close two parts of the hand separately to grasp and manipulate distinct objects consecutively. Fig. \ref{priority} (a) illustrates the open hand configuration at ${e_{g1}=e_{g2}=e_{m1}=e_{m2}=0}$ then the grasping components ${e_{g1}}$ and ${e_{g2}}$ are assigned with priority $\Upsilon = 0.5$ each. This allows the hand to close its two parts separately as shown in Fig. \ref{priority} (b) where the robot hand grasps the first object with little and ring fingers at ${e_{g1}=-0.04, e_{g2}= 0.05}$ and then it grips the second object in the tripod posture with ${e_{g1}=-0.11, e_{g2}= 0.22}$ as illustrated in Fig. \ref{priority} (c). Finally, the second object is rotated clockwise at $e_{m1}=-0.02$ to $0.11$, $e_{m2}= 0.25$ to $0.36$ such that it is manipulated without disturbing the pose of the first object as reported in Fig. \ref{priority} (d). Such characteristics of kernelized synergies help in performing different complex manipulation tasks such as writing with a pencil while holding a rubber, tightening a bolt while griping an extra nut, holding and pressing spray and many others. 
   
In task4, the robot is playing the board game \say{carrom} as demonstrated in Fig. \ref{carrom_game}. This task requires the precise grasping of the striker to move it to a desired position of attack and then perform the appropriate pushing motion, needed to guide the striker towards the goal pieces on the board. In Fig. \ref{carrom_game} (a), the robot hand grasps the striker by using grasping synergy coefficients ${e_{g1}=-0.24, e_{g2}= 0.18}$. It is then moved to the desired position on the board using the robot arm motion as illustrated in Fig. \ref{carrom_game} (b). The robot hand releases the striker at this desired location with ${e_{g1}=-0.19, e_{g2}= 0.14}$. The robot arm assumes the pose to bring the fingertips of the robot hand near the striker edge as displayed in Fig. \ref{carrom_game} (c). The robot hand first stretches in the fingers using manipulation coefficients in the range $e_{m1}=0.51$ to $0.59$, $e_{m2}= 0.05$ to $-0.08$ as represented in Fig. \ref{carrom_game} (d) and then stretches out its index, middle and thumb in the range of ${e_{m1}=0.1}$ to $-0.07$, $e_{m2}= -0.04$ to ${0.08}$ such that the striker gets pushed towards the pieces in the center as shown in Fig. \ref{carrom_game} (e). The surface of board is smooth and the friction between the striker and board is neglected, which is normally required in this game. The weight of striker is 15 grams, a force of 1.25 N, calculated using robot hand's motor currents, is applied to push the striker towards the center. The pushing motion in this task corresponds to the general postures taught to the robot in free hand configuration during training phase in Fig. \ref{objects} . Note that all the synergy coefficients exploited for different tasks are according to the training data set and can also be visualized from Fig. \ref{training_trajectory} for the different objects to be grasped and manipulated.

\section{CONCLUSIONS}

This research proposed a framework, called kernelized synergies to reuse the learned synergy subspace for precision grasping and dexterous manipulation of daily life objects. The postural synergies computed by tele-operating the robot hand for fundamental grasping and manipulation primitives were evolved over the duration of demonstrations to obtain the corresponding synergistic trajectories. The given synergistic trajectories were approximated with GMM-GMR in terms of Gaussian components to not only account for inconsistencies in the demonstrations but also to obtain a generalized probabilistic trajectory to reproduce the taught postures. In order to grasp and manipulate a new object, the KMP was exploited to deal with the environmental descriptors i.e, via-points and end points to adapt to the new shape and size respectively. The parametric synergistic trajectory was kernelized with KMP to preserve its grasping and manipulation characteristics so that the same synergistic subspace can be reused for new objects. The proposed framework was initially tested in the Syngrasp toolbox on two different simulated robot hand models and the stability of postures was evaluated and reported using force closure quality index. The lower values of quality index in case of anthropomorphic hand indicated its higher dexterity on stably grasping objects. For the experimental evaluation, four complex tasks similar to the daily life activities i.e, Pouring coffee and closing jar, opening toolbox latches, sequentially grasping and manipulating two objects, and playing carrom board game were performed and discussed. The results confirm that the proposed framework bridges the gap between the grasping and manipulation sub-spaces and is a hand agnostic approach provided the demonstration are carried out on the robot hand of interest.        

The possible extension to the proposed framework is the integration of visuo-tactile feedback for run-time adaption to the unknown and deformable objects. Moreover, the kernelized synergies can also be used for whole body motion planning in a reduced subspace by assigning priorities to the distinct tasks at different levels. Further, various kernels apart from frequently used Gaussian kernel can be exploited to investigate their effects on the performance of prehensile and non-prehensile manipulation tasks.

\end{document}